\documentclass{article}

\usepackage{arxiv}

\usepackage[utf8]{inputenc} % allow utf-8 input
\usepackage[T1]{fontenc}    % use 8-bit T1 fonts
\usepackage{hyperref}       % hyperlinks
\usepackage{url}            % simple URL typesetting
\usepackage{booktabs}       % professional-quality tables
\usepackage{amsfonts}       % blackboard math symbols
\usepackage{nicefrac}       % compact symbols for 1/2, etc.
\usepackage{microtype}      % microtypography
\usepackage{graphicx}
\usepackage{doi}
\usepackage{mathtools}
\graphicspath{ {./images/} }
\usepackage{wrapfig}
\usepackage[linesnumbered,ruled,vlined]{algorithm2e}
\usepackage{adjustbox}
\usepackage{caption}
\usepackage{bm}
\usepackage{xcolor}
\usepackage{threeparttable}

\title{Generative flow induced neural architecture search: Towards discovering optimal architecture in wavelet neural operator}

\date{} 					% Or removing it

\author{Hartej Soin \\
        Department of Applied Mechanics\\
        Indian Institute of Technology Delhi\\
        Hauz Khas, India 110016\\
        \texttt{am1210196@am.iitd.ac.in} \\
        \And
    	Tapas Tripura \\
        Department of Applied Mechanics\\
        Indian Institute of Technology Delhi\\
        Hauz Khas, India 110016\\
        \texttt{tapas.t@am.iitd.ac.in} \\
        \And
        Souvik Chakraborty \thanks{https://www.csccm.in/} \\
        Department of Applied Mechanics\\
        Yardi School of Artificial Intelligence (ScAI)\\
        Indian Institute of Technology Delhi\\
        Hauz Khas, India 110016\\
        \texttt{souvik@am.iitd.ac.in} 
}

\renewcommand{\headeright}{}
\renewcommand{\undertitle}{}
\renewcommand{\shorttitle}{Flow induced wavelet neural operator}

\hypersetup{
pdftitle={Flow induced wavelet neural operator},
pdfsubject={q-bio.NC, q-bio.QM},
pdfauthor={David S.~Hippocampus, Elias D.~Striatum},
pdfkeywords={First keyword, Second keyword, More},
}

\begin{document}
\maketitle

\begin{abstract}
We propose a generative flow-induced neural architecture search algorithm. The proposed approach devices simple feed-forward neural networks to learn stochastic policies to generate sequences of architecture hyperparameters such that the generated states are in proportion with the reward from the terminal state. 
% A flow-induced automated architecture generation framework is proposed for neural operators to learn discretization invariant solution operators of mechanics problems. Using the strategies from generative flow networks, we devise simple feed-forward neural networks to learn policies to generate sequences of architecture hyperparameters such that the generated states are in proportion with the reward from the terminal state.
We demonstrate the efficacy of the proposed search algorithm on the wavelet neural operator (WNO), where we learn a policy to generate a sequence of hyperparameters like wavelet basis and activation operators for wavelet integral blocks. While the trajectory of the generated wavelet basis and activation sequence is cast as flow, the policy is learned by minimizing the flow violation between each state in the trajectory and maximizing the reward from the terminal state. In the terminal state, we train WNO simultaneously to guide the search. We propose to use the exponent of the negative of the WNO loss on the validation dataset as the reward function. 
While the grid search-based neural architecture generation algorithms foresee every combination, the proposed framework generates the most probable sequence based on the positive reward from the terminal state, thereby reducing exploration time.
Compared to reinforcement learning schemes, where complete episodic training is required to get the reward, the proposed algorithm generates the hyperparameter trajectory sequentially. Through four fluid mechanics-oriented problems, we illustrate that the learned policies can sample the best-performing architecture of the neural operator, thereby improving the performance of the vanilla wavelet neural operator.
\end{abstract}

% keywords can be removed

\section{Introduction}
Since its first introduction around a couple of centuries ago, partial differential equations (PDEs) have remained an inextricable scientific tool for scientists and engineers in modeling natural phenomena like fluid flows, conduction, diffusion, weather forecasting, electrodynamics, and many more \cite{renardy2006introduction,sommerfeld1949partial}. 
Traditionally solved using methods such as the finite element and volume methods \cite{hughes2012finite,eymard2000finite}, which are highly mesh and resolution dependent, a recent push is seen towards employing discretization invariant alternatives like neural operators for solving these PDEs. 
Neural Operators learn the discretization invariant functional mappings between infinite-dimensional function spaces, acting as a generalization of neural networks that have been used to learn extremely complex functions. 
The diverse literature on neural operators includes the universal approximation theorem \cite{chen1995universal} based Deep Operator Network (DeepONet) \cite{lu2019deeponet} and physics-informed DeepONet \cite{wang2021learning}, graph discretization-based Graph Neural Operator (GNO) \cite{li2020neural}, spectral convolution-based Fourier Neural Operator (FNO) \cite{li2020fourier}, Wavelet Neural Operator (WNO) \cite{TRIPURA2023115783}, and physics-informed WNO (PIWNO) \cite{navaneeth2024physics}.
While DeepONet is developed over the feed-forward neural networks, FNO and WNO use convolution operations to parameterize the neural networks in the feature space. While both FNO and WNO remained discretization invariant, wavelets use both frequency and spatial information to learn the features effectively. 
% In particular, given a function space containing PDE inputs like source terms, the initial and boundary conditions, and the geometry information, a neural operator learns functional mapping, which maps a set of inputs from the given input function space to a unique function in the corresponding solution space. This also makes the neural operator's discretization invariant, i.e., once trained, it may be used for predicting solutions for different sets of input functions at different spatial locations, as opposed to the neural networks that can take only finite dimensional vector inputs and can predict only at training locations. 
Recent works in neural operators also include nonlinear manifold decoder (NOMAD) \cite{seidman2022nomad}, Laplace neural operator (LNO) \cite{cao2023lno}, and aliasing improved Spectral Neural Operator (SNO) \cite{fanaskov2023spectral}. Nevertheless, all the neural operator frameworks consist of an additional set of hyperparameters on top of the standard neural networks hyperparameters, like the latent dimension of branch and trunk net in DeepONet, Fourier mode number in FNO, and choice of wavelet basis in WNO, which makes tuning more involved and cumbersome. To that end, we propose a generative flow-induced neural architecture search algorithm tailored towards neural operators for automated hyperparameter selection and efficient learning of discretization invariant solution operators of parametric PDEs. 

The main aim of automated hyperparameter optimization in representation learning is to automatically learn the hyperparameters of deep neural networks to infer minimal bias in the trained network. A brief survey on the algorithm proposed for neural architecture search (NAS) in deep learning can be found in \cite{wistuba2019survey,elsken2019neural,10.1145/3447582,liu2021survey}. 
Naive architecture generation algorithms like grid search perform experiments over all possible combinations of hyperparameter space. In the context of neural operators, which are trained over a family of PDEs, performing such a brute-force architecture search requires a humongous amount of computational power and time. 
Efficient Neural Architecture Search (ENAS) \cite{pmlr-v80-pham18a}, on the other hand, improves the search time by incorporating a strategy for parameter sharing across different neural architectures. 
Methods like Progressive Neural Architecture Search (PNAS) use sequential model-based optimization (SMBO) to perform hyperparameter search based on the increasing complexity of generated architectures \cite{liu2018progressive}. Despite success, these strategies are limited to convolutional neural networks. 
Methods involving reinforcement learning for NAS are also proposed, which use expected accuracy over the validation dataset as a reward for automating the generation of novel neural network architectures. The reinforcement learning (RL) based methods for NAS like BlockQNN \cite{8578355}, MetaQNN \cite{baker2017designing}, and NAS-RL \cite{zoph2017neural} have been shown to be particularly successful. 
% However, in neural operators, training over a family of PDEs takes considerable computing time. 
%The performance of the neural operators using the RL-based strategies may only be judged after this training before we associate a cost with the architecture. In such a case, the information on the optimal performance over sufficient iterations may be lost with typical RL-based methods due to varying convergence rates for different architectures.
In RL-based methods, the objective is generally directed towards actions that maximize the reward. In this work, our objective is to sample the trajectories proportional to the distribution of the reward function relating to the cost and to exploit this capability to produce a diverse set of architectures requiring only partially trained neural operators, a subset of which may then be trained to convergence.
To solve this, an elegant architecture search method capable of generating a diverse set of architectures, unlike the deterministic RL-based approaches, is required. 

A similar strategy exists in the generative flow networks (GFlowNet) \cite{NEURIPS2021_e614f646,10.5555/3648699.3648909}, where the neural network agents learn a stochastic policy from a sequence of actions to generate a compositional object such that the objects are generated in proportional to the reward of the object. In GFlownet, each state in the compositional object is built sequentially, with a reward assigned to it, like in RL. However, the reward is estimated only at the terminal state, i.e., when an object is created. The policy converges when the total discrepancy between in- and out-flow from states in the generated sequence vanishes, i.e., when the incoming and outgoing flow into and out of each state match and the reward is maximized. Once training succeeds, the learned policy takes an action with probability proportional to the outgoing flow. 
A brief study on the successful application of GFlowNets as a generative model can be observed in the generation of new samples of molecules \cite{NEURIPS2021_e614f646} and active learning \cite{pmlr-v162-jain22a,pmlr-v202-jain23a}. 
In this work, we utilize the concept of turning rewards into a generative policy from GFlowNet to devise a neural architecture search algorithm for the neural operators. 
The probabilistic policy is learned by using a flow-consistent loss function, which takes into account the discrepancy in the flow between the states of the generated sequence of network hyperparameters and prediction reward from the underlying neural operator. The proposed framework consists of a series of neural networks, each trained to sample a hyperparameter, except the terminal network, which is set as the underlying neural operator. 
The series of networks learn the probability of taking action given the previous state, which is referred to as the flow associated with that state. The total flow into the network is estimated as the sum of the in- and out-flow from the states. The terminal network uses the generated hyperparameters from previous networks to return a reward. 
The reward is estimated at the end of the terminal neural operator after having generated the solution of the underlying PDE, like the episodic setting of conventional RL, which, in this case, is a function of the prediction loss. 

Since the first introduction, WNO has gone through a significant number of independent developments, including Waveformer \cite{N2024111253} for long-term prediction, wavelet elastography for medical imaging \cite{tripura2023elastography}, multi-fidelity WNO (MFWNO) \cite{thakur2022multi} for learning from multi-fidelity dataset, neural combinatorial WNO (NCWNO) for multi-physics and continual learning of solution operators \cite{tripura2023foundational}, generative adversarial WNO (GAWNO) for generative modeling \cite{rani2024generative}, and differentiable physics augmented WNO (DPA-WNO) \cite{chakraborty2023dpa} as a deep physics corrector. 
All these frameworks use wavelet integral blocks, where wavelet decomposition is used to project the features to a space-frequency localized space and subsequently convoluted with neural network kernels to learn the features from data, followed by a non-linear activation operator. 
In addition to the standard neural network parameters like the number of hidden layers, channel dimension, number of epochs, learning rate, batch size, and regularizer, these integral blocks introduce additional parameters like the choice of wavelet basis, wavelet decomposition level, and activation operators. Choosing the right combination of these wavelet hyperparameters for each wavelet integral block requires prior experience, which further becomes cumbersome as the number of wavelet integral blocks increases. 
Therefore, the large hyperparameter space makes WNO a natural choice for illustrating the developed neural architecture search algorithm. 
% To alleviate this problem, we use the reward-based policy learning strategy from generative flow networks to learn stochastic policy to generate a sequence of wavelet basis and activation operators for the flow-induced WNO (Flow-WNO). 
% GFlowNets produce a diverse set of architectures with an attempt to make probabilities of the samples proportional to the rewards defined over partially trained WNOs. From these diverse sets of candidates, we choose $k$ best-performing architectures, which are trained fully to decide on the optimal architecture.

The main contribution of the proposed framework can be encapsulated into the following points:
\begin{itemize}
    \item An automated neural architecture search algorithm rooted in GFlowNet is proposed for neural operators. The efficacy is exemplified in WNO by learning a stochastic policy to generate a sequence of wavelet basis and activation operators. 
    \item By satisfying the in- and out-flow from terminal states, the proposed framework achieves optimality criteria over the neural operator architecture.
    \item The proposed framework learns to generate the network architecture by learning a probabilistic policy instead of a grid search over the parameter space, thereby reducing tuning time. In the reward-based policy learning setup, the final compositional sequence is generated sequentially based on partially trained WNO, which further reduces the computational costs.
    \item The proposed framework preserves all the benefits of the parent neural operator architecture; in this case, discretization invariant operator learning of a family of parametric PDEs. 
\end{itemize}

The rest of the paper is arranged as follows. Section \ref{methodology} gives a brief mathematical introduction to WNO and GFlowNet. Section \ref{sec:proposed} briefly illustrates the proposed framework on WNO. Section \ref{results} consists of the benchmark examples over which we test our proposed method, including the Burger, Darcy, and Navier Stokes equations. Section \ref{sec:conclusion} concludes this study by reviewing the features of the proposed architecture generative framework.

\section{Background on wavelet neural operator and flow networks}\label{methodology}
% We will begin by giving an overview of WNOs and GFlowNets, followed by a deep dive into how we utilize GFlowNets to come up with WNO architectures in this paper.
A short overview of WNO and GFlowNet is presented in this section, which will be used in the next section to construct the proposed framework of flow-induced wavelet neural operator (FWNO). 

\subsection{\textbf{Wavelet Neural Operator (WNO)}}\label{WNO}
Neural operators are a class of deep learning algorithms that learn the functional mapping between two infinite-dimensional function spaces, the input and output spaces, as opposed to the artificial neural networks (ANN) that learn the function map between two finite-dimensional vector spaces. Talking specifically in terms of PDEs, neural operators learn the mapping from the input space, comprised of the initial and boundary conditions, geometry, physical parameters, and source term to the solution space of the PDE. Thus, after training, a neural operator can be used to predict the solution of a family of PDEs, whereas the ANNs need retraining for every input combination. Wavelet Neural Operator (WNO) is one such deep neural operator, which uses the frequency-spatial localization property of wavelet transformation to learn the feature space in wavelet space. 

For a mathematical representation, we consider the $n$ dimensional fixed domain $D\in \mathbb{R}^n$, bounded by $\partial D$. Over the domain $D$, we consider the Banach spaces $\mathcal{A}:=C(D;\mathbb{R}^{d_a})$ and $\mathcal{U}:=C(D;\mathbb{R}^{d_u})$ such that $\lambda \in \mathbb{R}^{d_a}$ and ${u} \in \mathbb{R}^{d_u}$ are the input and outputs in the Banach spaces $\mathcal{A}$ and $\mathcal{U}$, respectively. Between the spaces $\mathcal{A}$ and $\mathcal{U}$, we define the nonlinear PDE operator $\mathcal{N}: \mathcal{A} \ni \lambda \mapsto {u} \in \mathcal{U}$, which maps a given set of input parameters $\lambda$ to a unique solution ${u}$.

Given an $N_s$ number of input-output training pairs $\{(\lambda_1,{u}_1), (\lambda_2,{u}_2), \ldots, (\lambda_{N_s},{u}_{N_s})\}$ are available such that ${u}_j=\mathcal{N}(\lambda_j)$, WNO aims to approximate $\mathcal{N}$ through a parameter space $\Omega$, i.e., $\mathcal{N}: \mathcal{A} \times \Omega \mapsto \mathcal{U}$, where $\Omega$ denotes the finite-dimensional parameter space for the neural network. 
The $m$ point discretization of the domain $D$ yields the set $\{\bm{\lambda}_j\in \mathbb{R}^{m \times d_a},\bm{u}_j\in \mathbb{R}^{m\times d_u}\}, j=1 \ldots, N_s$. To increase the channel depth of the parameter space $\Omega$, WNO raises the inputs $\lambda(x) \in \mathbb{R}^{d_a}$ to a high dimensional space $\mathbb{R}^{d_a}$ through a local transformation $\mathcal{P}(\lambda(x)):\mathbb{R}^{d_a}\to \mathbb{R}^{d_v}$, denoting it as $v_0(x) \in \mathbb{R}^{d_v}$. This can be achieved using either a fully connected neural network (FNN) or a $1\times 1$ convolution (CNN). The uplifted inputs are passed through a series of recursive wavelet integral blocks $\mathcal{G}:\mathbb{R}^{d_v} \mapsto \mathbb{R}^{d_v}$. The updates $v(j+1)=\mathcal{G}(v(j))$ via wavelet integral blocks $\mathcal{G}$ are defined as,
\begin{equation}
    v_{j+1}(x) := g((\mathcal{K}(\lambda;\omega)*v_j(x)+Wv_j(x)) \;\; x\in D, j\in [1,l]
\end{equation}
where $g(x):\mathbb{R}\to \mathbb{R}$ is a non-linear activation function, $W:\mathbb{R}^{d_v}\to \mathbb{R}^{d_v}$ is a linear transformation, $*$ represents the convolution operation and $\mathcal{K}:\mathcal{A} \times \Omega \mapsto \mathcal{U}$ is the integral operator over $C(D;\mathbb{R}^{d_v})$. Since we use a neural network architecture, we can represent the integral operator $\mathcal{K}(a;\omega)$ as a kernel integral operator with parameter $\omega\in\Omega$.
These wavelet integral blocks extract relevant feature maps from the data through convolution in the wavelet domain. In the end, a local transformation $\mathcal{Q}(v(x)):\mathbb{R}^{d_v} \mapsto \mathbb{R}^{d_u}$ such that $u(x)=\mathcal{Q}(v_l(x))$ is used to reduce the channel depth to the desired solution space. 
Using the concept of element-wise multiplication in spectral space, the convolution in kernel integration is performed in the wavelet space. Parametrizing the kernel in the wavelet domain to learn the kernel leads to the WNO framework. For a brief review, readers are referred to \cite{TRIPURA2023115783,navaneeth2024physics}.

\subsection{Generative Flow Network}
GFlowNets are a class of generative models that aim to generate a compositional object $x \in \mathcal{X}$, an object that may be represented as a sequence of discrete actions applied iteratively to a base state. At each stage of the iterative updates, we get a partially constructed object which is sequentially updated based on the actions from the remaining iterations. It may be represented as a directed acyclic graph (DAG) of a Markov Decision Process (MDP), $\mathcal{D}=(S,A)$ where $S$ is the set of states possible, or nodes in the graph where $\mathcal{X} \subset S$, and $A \subseteq S\times S$ is the set of all possible actions or the directed edges of DAG. We further define $A(s)\subseteq A$ as the set of actions allowed at state $s$, and $A'(s)$ as the set of sequences of all actions allowed after state $s$. We call state $p$ the parent of state $s$ if the edge $(p\to s)\in A$ and state $c\in S$ a child of $s$ if edge $(s\to c)\in A$. The complete trajectory is defined as $\tau =(s_0,s_1,..,s_T)$ where $s_j\in S$ is the $j^{th}$ state, $s_0$ is the base state and $s_T=x\in \mathcal{X}$ is the terminal state.

The reward for any state $s\in S$ is denoted as $R(s)$, where the reward is zero for all non-terminal (intermediate) states. The reward is independent of intermediate states for all terminal states $x$. To that end, the flow network is resented as an MDP with the root node as the source $s_0$, the in-flow as $Z$, and the out-flow as $R(x')$ that flows out of each terminal state $x'$. Any state $s$, when subjected to action $a$, leads to a new state $s' \in S$, represented as $\mathcal{T}(s,a)=s'$. $F(s,a)$ is the flow from state $s$ to $s'$ and $F(s)$ is the total flow out of state $s$. A policy $\pi(a \vert s)$ decides the flow through the sequence $\tau$. The policy $\pi(a \vert s)$ sequentially builds the compositional object $x$ with probability $\pi(x)$. This is defined as,
\begin{equation}
    \pi(a \vert s)=\frac{F(s,a)}{F(s)} ,
\end{equation}
where $\pi(x)= R(x)/Z$ and $Z=\sum_{x'\in X}R(x')$. From the definition, it is understood that $\pi(x)$ is proportional to $R(x)$, i.e., the policy generates $x$ with a probability proportional to the associated reward.
Learning the policy $\pi(x)$, therefore, requires minimizing the imbalance between the incoming flow and outgoing flow at a node $s \in S$ and maximizing the reward $R(x)$. The flow conditions in a flow network as well as the reward function, can incorporated into the following flow consistency equation, 
\begin{equation}\label{eq:loss}
    \sum_{s,a:\mathcal{T}(s,a)=s'} F(s,a)=R(s')+\sum_{a'\in A(s')}F(s',a') .
\end{equation}
Note that in this equation $R(s')=0$ for intermediate states $s'$ while for terminal state $s'=x\in \mathcal{X}$, the outflow is zero, i.e., $\sum_{a' \in A(s')}F(s',a')=0$, since $A(s')$ is an empty set.
The minimization between the left and right-hand sides of the Eq. \eqref{eq:loss} becomes the loss function of GFlowNet. 
Since the flow values at the root nodes are exponentially larger than the nodes at the later stages of the flow network, the gradient weights for smaller predictions pose numerical issues to the learning of the neural network, due to which the logarithm of inflow and outflow from a node is matched. 
Finally to approximate the policy $\pi(a \vert s)$, GFlowNet minimizes the following log-scale objective function,
\begin{equation}\label{eq:logloss}
    \mathcal{L}(\tau) =\sum_{s'\in \tau\neq s_0}\left[\log\left\{\sum_{\substack{s,a,\mathcal{T}(s,a)=s'}}\exp\left(F^{\log}\left(s,a\right)\right)\right\} -\log\left\{R(s')+\sum_{a'\in A(s')}\exp\left(F^{\log}\left(s',a'\right)\right)\right\}\right] ,
\end{equation}
where $F^{\log}(s,a)=\log F(s,a)$. The loss function $\mathcal{L}(\tau)$ minimizes the imbalance between inflow and outflow over a trajectory $\tau$. In a similar manner to the MDP, where the temporal difference between successive states is used to update the value function of states using the Bellman equation, by minimizing $\mathcal{L}(\tau)$, we learn the policy $\pi(a \vert s)$ to take best-performing actions. 

\begin{figure}[!b]
    \centering
    \includegraphics[width=\textwidth]{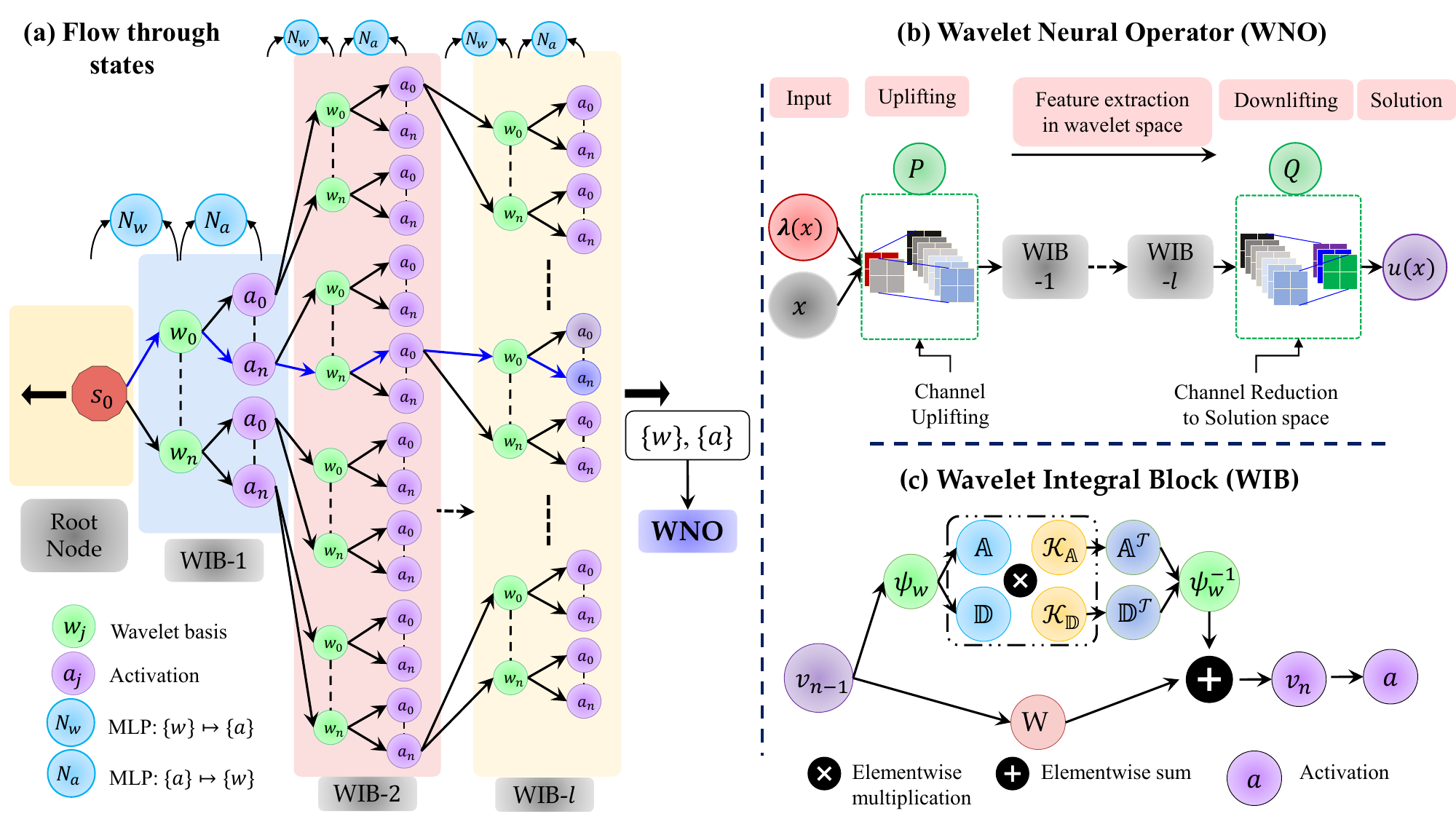}
    \caption{Schematic architecture of the proposed flow-induced wavelet neural operator. (a) The flow network of wavelets and activation operators leverages the DAG structure of the Markov decision process. The algorithm chooses a binary set consisting of wavelet basis and activation function for each of the wavelet integral blocks (WIBs). The arrows leading from one state to the next represent the flow. Starting at $s_0$, we sample the next action (wavelet/activation) proportional to the output of the Neural Networks $N_w$/$N_a$, which is the flow from one node to the next nodes. A representative set of states chosen by FWNO with maximum probability is shown as $\{w_0,a_n,w_n,a_0,\ldots,w_0,a_n\}$ given the highest flow corresponding to trajectories indicated by the blue path. The loss $\mathcal{L}$ over the trajectory is accumulated, and the agents $N_w$ and $N_a$ are updated at the end of the trajectory. (b) The schematic of the wavelet neural operator. It uses the selected set of $\{w,a\}$ to parameterize the WNO kernels in wavelet space. (c) The wavelet integral block uses selected basis $w_i,i=1,\ldots,\ell$ to perform forward and inverse wavelet transforms $\psi_{w}$ and $\psi_{w}^{-1}$. The approximated $\mathbb{A}$ and detailed $\mathbb{D}$ features of the inputs are convolved with the kernels $\mathcal{K}_{\mathbb{A}}$ and $\mathcal{K}_{\mathbb{D}}$.}
    % \caption{Schematic architecture of the proposed flow-induced wavelet neural operator exemplified on a WNO with 2 wavelet kernel integral layers. The algorithm chooses a binary set consisting of wavelet basis and activation function for each of the wavelet integral layers. The arrows leading from one state to the next represent the flow in terms of transparency. Starting at $s_0$, we sample the next action(basis/activation) proportional to the output of the Neural Network ($N_w$/$N_a)$, which is the flow from one node to the next nodes. The GFlowNet chooses the architectures $\{w_0,a_0,w_1,a_0\}$ and $\{w_0,a_0,w_1,a_1\}$ with maximum probability given the highest flow corresponding to trajectories leading to these states. The loss $\mathcal{L}$ over the trajectory is accumulated, and the agents $N_w$ and $N_a$ are updated at the end of the trajectory.}
    \label{fig:method}
\end{figure}

\section{Flow induced wavelet neural operator}\label{sec:proposed}
This section instantiates the flow-induced wavelet neural operator (FWNO) by incorporating the reward-based policy learning mechanism inside the vanilla WNO. The concept of flow is used to construct a compositional object containing binary sets of an activation operator and a wavelet basis. Each binary set is used for wavelet decomposition and nonlinear activation during the kernel parametrization in each wavelet integral block of the WNO architecture. By learning to generate the sequences of binary sets in proportion to the reward based on the prediction error, an optimal operator learning architecture with improved accuracy is achieved over the vanilla WNO. 

For generating the best-performing sequence of wavelet basis functions and activation operators, we propose the \emph{reward function} as the \emph{validation loss for the WNO architecture}, i.e., $R(s_T) \propto \mathcal{L}(u_{\text{Val}},u_{\text{Val}}^{*})(x)$ such that $u_{\text{Val}}^{*} = \mathcal{N}_{\Omega}(\lambda_{\text{Val}}; \tau)(x)$, where $\lambda_{\text{Val}}$ is the unseen input parametric field, $u_{\text{Val}}$ and $u_{\text{Val}}^{*}$ are the true and predicted solution of the true PDE operator $\mathcal{N}$, and $\mathcal{N}_{\Omega}$ is the WNO. The trajectory $\tau$ contains the set of wavelet basis and activation operators. 
In a physics-informed neural operator framework, this reward function can be tuned depending on the requirement, like giving more weight to the boundary conditions, the physics-informed constraints, smoother curves, and other such criteria. In our data-driven framework, we will consider the $L^2$ relative norm of the prediction error. 

Starting at empty state $s_0 = \{\cdot, \cdot \}$, we generate the WNO model $\mathcal{M}$. The model $\mathcal{M}$ is then trained on the paired dataset $\{(\lambda_1,{u}_1), (\lambda_2,{u}_2), \ldots, (\lambda_{N_s},{u}_{N_s})\}$ to learn the operator $\mathcal{N}_{\Omega}$. The operator $\mathcal{N}_{\Omega}$ returns a validation loss $\mathcal{L}(u_{\text{Val}},u_{\text{Val}}^{*})$ over the paired validation dataset $\{\lambda_{\text{Val}},u_{\text{Val}}\}$. The reward $R(s_T) = \exp(- \mathcal{L}(u_{\text{Val}},u_{\text{Val}}^{*})(x))$ for the model is then computed as the exponent of the inverse of this loss to ensure a non-negative reward. 
Let $\mathcal{W} =\{w_1, w_2, \ldots, w_{\ell}\}$ and $A = \{a_1, a_2, \ldots, a_{\ell}\}$ denote the sets of wavelet basis functions and activation functions in our operator framework, where each pair $\{w_j,a_j\}$ is associated with $j^{th}$ wavelet integral block. Two neural networks $N_w(s;\theta)$ and $N_a(s;\phi)$ are used to learn distributions of the flow across the state $s \in S$. Here $N_w(s;\theta)$ is used to construct the trajectory of wavelet bases, and $N_a(s;\phi)$ is used to construct the trajectory of activation operators in the wavelet integral blocks. 
At each state $s$, the difference between the total inflow $Z$ to $s$ and the total outflow from $s$, which is $R$ for the terminal state, are then minimized over the trajectory to train models $N_w$ and $N_a$. The selection of a binary set of wavelet basis and activation function from the current state $s\in S$ using the neural networks $N_w$ and $N_a$ can be represented using the sequential graph as follows, 
\begin{equation}
    \tau : s \xrightarrow{\text{$N_w$}} \{w\} \xrightarrow{\text{$N$}_a} \{w,a\} , 
\end{equation}
where $w \in \mathcal{W}$ and $a \in A$ are sequentially selected based on outputs of the two networks $N_w$ ad $N_a$. This partial set is further constructed by minimizing the discrepancy between the in-and-out flow over the other wavelet integral layers, which is represented as,
\begin{equation}
    s = s \cup \{w,a\}.
\end{equation}
Combining $n$ such wavelet layers gives us the final compositional object comprising of the incremental sets of wavelet basis and activation function for each of the layers, represented as the trajectory, 
\begin{equation}\label{eq:trajectory}
    \tau : s_0 \xrightarrow{\text{$N_\theta$}} \{w_1\} \xrightarrow{\text{$N$}_\phi} \{w_1,a_1\} \xrightarrow{\text{$N$}_\theta} \{w_1,a_1,w_2\} \cdots\xrightarrow{\text{$N_{\phi}$}} \{w_1,a_1,w_2,a_2,\cdots w_n,a_n\} .
\end{equation}
Using the compositional set of wavelet bases and activation functions, the WNO model $\mathcal{M}: \mathbb{R}^{d_a}\to\mathbb{R}^{d_u}$ is trained to learn the operator $\mathcal{N}: \mathcal{A} \times \Omega \mapsto \mathcal{U}$ with input $\lambda \in \mathcal{A}$ and output $u \in \mathcal{U}$. The complete process of learning the operator can be represented as an iterative process between the input and output can be represented as, 
\begin{equation}
    u(x) = \mathcal{Q}(a_{\ell}(\mathcal{G}_{\ell}\cdots(a_2(\mathcal{G}_{1}(a_1(\mathcal{G}_0(\mathcal{P}(\lambda);w_1));w_2)\cdots; w_{\ell})))(x),
\end{equation}
where the transformations $\mathcal{P}: \mathbb{R}^{d_a} \to \mathbb{R}^{d_v}$ and $Q:\mathcal{R}^{d_v} \to \mathbb{R}^{d_u}$ are defined in Section \ref{WNO}. Similarly the iterations $\mathcal{G}:\mathbb{R}^{d_v} \to \mathbb{R}^{d_v}$ are represented as,
\begin{equation}
    \mathcal{G}_{j}(a_{j},w_{j}) := \mathcal{K}(\lambda; a_{j} \in A, w_{j} \in \mathcal{W}, \omega \in \Omega) \cdot v_{j}(x) + W \cdot v_j(x), \;\; j \in [1, \ell]
\end{equation}
where $W(\cdot)$ and $K(\cdot, \cdot)$ are defined in the Section \ref{WNO} conditioned over the activation operator $a_j$ and wavelet basis function $w_j$ given the network parameters $\Omega$. To calculate the reward, we consider the loss $\mathcal{L}$ as the imbalance in inflow and outflow for each state in the flow network over the trajectory $\tau$ as,
\begin{equation}\label{eq11}
    \mathcal{L}(\tau) = \sum_{\substack{s_j\in \tau\neq s_0, \\ \mathcal{T}(s_{j-1},a)=s_j}} \left[F(s_{j-1},a)-\sum_{a'\in A}F(s_j,a')-R(s_j)\right]^2 . 
\end{equation}
During training, the loss $\mathcal{L}$ becomes the objective function for tuning the parameters of the networks $N_w(\cdot; \theta)$ and $N_a(\cdot;\phi)$. The training process may then be represented as,
\begin{equation}
    \{w_1,a_1,w_2,a_2,\cdots w_n,a_n\} \xrightarrow{\mathcal{D}}R(s_j) \xrightarrow{\mathcal{L}(\tau)}\theta,\phi=\underset{\theta,\phi}{\arg\min}\{\mathcal{L}\}
\end{equation}

For ease of understanding, a schematic representation of the proposed flow-induced wavelet neural operator is illustrated in Figure \ref{fig:method}. 
\begin{algorithm}[t]
    \caption{GFN-WNO}
    \label{algo}
    \KwData{Input dataset $\mathcal{D}$, set of wavelets $\mathcal{W}$, set of non-linear activation operators $\mathcal{A}$, number of wavelet integral blocks $\mathcal{G}$.}
    %\KwResult{Output models $N_1$ and $N_2$ to select basis and activation functions}
    \textbf{Initialisation:} The empty state $s_0$, fully connected neural networks $N_w$ and model $N_a$, Loss function $\mathcal{L}$. \\
    % \textbf{Procedure:}
    \For{$i \leftarrow 1$ \KwTo \text{iteration}}{
        \For{$j \leftarrow 0$ \KwTo $2n-1$}{
            \If{$j|2$}{
                $w\leftarrow  Categorical(N_w(s))^*$ \\
                Inflow $F_i\leftarrow N_w(s)[w]$ \\
                add $w$ to $s$ \\
                Outflow $F_o\leftarrow \sum_a N_a(s)[a]$
            }
            \Else{
                $a\leftarrow Categorical(N_a(s))^*$ \\
                Inflow $F_i\leftarrow N_a(s)[a]$ \\
                add $a$ to $s$ \\                
                \If{$j\neq 2m-1$}{
                    Outflow $F_o\leftarrow \sum_w N_w(s)[w]$ \\
                    Reward $R\leftarrow 0$
                }
                \Else{
                    Outflow $F_o\leftarrow 0$\\
                    Train WNO model $\mathcal{M}$ with architecture $s$ \\
                    Reward $R\leftarrow$ exp(-(Test loss for the model $\mathcal{M}$))
                }
            }
            $\mathcal{L} \leftarrow \mathcal{L} + (F_i-F_o-R)^2$ 
        }
        Train $N_w$ and $N_a$ over objective $\mathcal{L} \leftarrow 0$
    }
    \Return{$N_w$ and $N_a$}
\end{algorithm}

\subsection{Initial setup}
We instantiate the FWNO framework by selecting a sufficiently rich set of wavelets $\mathcal{W}$ and non-linear activation operators $A$. For the purpose of learning a policy to sample states of the wavelets and activation operators in the sets $\mathcal{W}$ and $A$, two neural networks $N_w$ and $N_a$ are set up as feed-forward neural networks. Here, a state $s \in S$ is described as the sequence of basis and activation functions in the trajectory in \eqref{eq:trajectory}. The base state $s_0$ here is an empty set. The terminal state is denoted as $s_{2n}$, where $n$ is the number of wavelet integral blocks $\mathcal{G}$ needed for satisfactory convergence. 
The FWNO architecture is then generated sequentially by selecting a wavelet cum activation pair for each of the wavelet integral blocks based on the policy from $N_w$ and $N_a$. During the architecture generation process, the current states are alternatively passed through $N_w$ and $N_a$ to probabilistically sample the wavelets and activation operators for the current state (see section \ref{sec:selection}). At the end of the policy learning, the terminal state $s_{2n} = \{w_1,a_1,\ldots,w_n,a_n\}$ provides us with the best-performing FWNO architecture, where $w_j \in \mathcal{W}$ is the wavelet basis function and $a_j \in A$ is the non-linear activation operator for $j^{th}$ wavelet integral block and the state space $S$ contains $n$ such pairs of wavelets and activation operators.

\subsection{Sequential construction of the architecture}\label{sec:selection}
At the beginning the initial state $s_0$ is passed to the $N_w$ to get a distribution $\pi_1^{\{w\}}$ over $\mathcal{W}$. We sample the set $\mathcal{W}$ with a probability $\pi^{\{w\}}(s_0) \propto \pi^{\{w\}}_1$ to get the first state $s_1 = \{w_1\}$. The sampled state $s_1$ is then passed into the network $N_a$ to get a distribution $\pi_1^{\{a\}}$ over the activation operator space $A$. On sampling from the set $A$ with a probability $\pi^{\{a\}}(s_1) \propto \pi^{\{a\}}_1$ we get the second state $s_1 = \{w_1,a_1\}$. This wavelet-activation pair constructs the first wavelet integral block $\mathcal{G}_1$. 
This state is sequentially passed to $N_w$ and $N_a$ till the terminal state $s_{2n}$ is reached and all the wavelet and activation operator pairs are sampled with the highest probability. Note that we are training two neural networks $N_w$ and $N_a$ here and states of the form $s_{2r}$ are the inputs for $N_w$ and states of the form $s_{2r+1}$ are the inputs for $N_a$, such that,
\begin{equation}
    \begin{aligned}
        s_{2r} &= \{s_{2r-1},w\sim N_w(s_{2r-1}))\} \\
        s_{2r+1} &= \{s_{2r},a\sim N_a(s_{2r})\}
    \end{aligned}
\end{equation}

\subsection{Training the networks to learn flow}
A trajectory is considered complete on reaching the terminal state $s_{2n} = \{w_1,a_1,\ldots,w_n,a_n\}$. For a trajectory $\tau$ of the form in Eq. \eqref{eq:trajectory} we first calculate the inflow $F(s_{j-1},a)$ from state $s_{j-1}$ to $s_j, 0<j \leq n$. 
As defined before, we continue to use the notation $\mathcal{T}(s_{j-1},a) = s_j$ to denote the result of action $a$ on state $s_{j-1}$ leading to state $s_j$. 
Then we find the total outflow from $s_j$ which is given as $\sum_{a'\in A(s(j))}F(s_j,a')$ where $A(s) \subseteq A$ is the set of all actions possible on state $s$. 
We also calculate the reward for $s_j$ given as $R(s_j)$, which is the inverse of the exponential of training loss for the WNO architecture given by $s_j$ if it is a terminal state and 0 otherwise. 
By including the inflow, outflow, and reward from WNO, we minimize the discrepancy in the information flow between the states. The objective function for a trajectory $\tau$ that is to be minimized is given as \ref{eq11}. The flow $F(s_{j-1},a)$ is parametrized by the neural network $N_w$ if $j$ is odd and by the neural network $N_a$ otherwise.
% and is given by the $a_{th}$ entry of the output of the neural network. 
Therefore, minimization of the objective function optimizes the network parameters of $N_w$ and $N_a$ while learning the policy to select the best-performing sequence of wavelets and activation operators. 
For ease of implementation, the algorithm of the proposed framework is given in Algorithm \ref{algo}, which briefly illustrates the implementation steps of the proposed FWNO.

\section{Numerical Results}\label{results}
In this section, we illustrate the performance of the proposed framework on four mechanics examples. The experiments include standard benchmarks like the 1D Burgers equation, the 2D Darcy flow equation, and a 2D incompressible Navier-Stokes equation. As previously stated, we illustrate the efficacy of the proposed generative flow-induced neural architecture search algorithm in learning the optimal neural architecture of WNO. The resulting WNO architecture is referred to as Flow-induced WNO (FWNO). The improvement achieved is reported against the vanilla WNO architecture-based results reported in \cite{TRIPURA2023115783}. 
During the training of FWNO, we calculate the reward for each generated architecture over 100 epochs only. The optimal architecture obtained is then trained for 500 epochs to obtain the optimal solution. 
% The hyperparameter details are provided in the Appendix \ref{sec:hyperparameter}.

\paragraph{Hyperparameter settings.} 
The fully connected neural networks denoted as $N_w$ and $N_a$, share identical hyperparameters across all cases except for the Darcy flow equation in a rectangular grid. For the former cases, these hyperparameters consist of a single hidden layer comprising 16 nodes, followed by the Leaky ReLU activation function. However, the Darcy equation deviates from this norm, featuring a hidden layer with 128 nodes instead.
Both $N_w$ and $N_a$ undergo optimization using the Adam Optimizer, employing a fixed learning rate of $10^{-3}$. The training process spans 500 iterations for all scenarios except the Navier-Stokes equation, where it is limited to 100 iterations. Table \ref{table:hyparams} specifies the specific hyperparameters employed in the training of FWNO architectures across the various cases taken from \cite{TRIPURA2023115783}.

\begin{table}[!ht]
    \centering
    \caption{Hyperparameters details for example problems}
    \label{table:hyparams}
    \small % Smaller font size for column names
    \begin{tabular}{lcccccc}
        \toprule
        Example & Levels of Wavelet & Batch Size & Learning & \multicolumn{2}{c}{Scheduler (Learning Rate)} & Weight \\
        \cmidrule(lr){5-6}
        & Decomposition & & Rate & Step Size & $\gamma$ & Decay \\
        \midrule
        Burgers' diffusion dynamics & 8 & 10 & $10^{-3}$ & 100 & 0.5 & $10^{-4}$ \\
        Darcy equation (square domain) & 4  & 20 & $10^{-3}$ & 50 & 0.75     & $10^{-4}$ \\
        Darcy equation (with notch)& 3 & 25 & $10^{-3}$ & 50 & 0.75 & $10^{-4}$ \\
        Navier-Stokes equation & 3 & 20 & $10^{-3}$ & 50 & 0.75 & $10^{-4}$ \\
        \bottomrule
    \end{tabular}
\end{table}

\subsection{Burgers diffusion dynamics}
The first example we consider for the numerical illustration is the 1D Burgers equation used in modeling flow in fluid mechanics, traffic flow, and acoustics. The following parabolic partial differential equation describes the Burgers diffusion dynamics,
\begin{equation}
    \begin{aligned}
        \frac{\partial u}{\partial t} + \frac{1}{2}\frac{\partial u^2}{\partial x} &= \nu\frac{\partial^2 u}{\partial x^2},  \;\; x \in [0,1], t \in [0,1] \\
        u(x=0,t) &= u(x=1,t) \\
        u(x,t=0) &= u_0(x)\\
    \end{aligned}
\end{equation}
where we consider periodic boundary conditions. In the above equation, $\nu \in \mathbb{R}^+$ defines the viscosity of the flow. The training dataset is generated for different initial conditions $u_0(x)$, where $u_0(x)$ is modeled as a Gaussian Random field $u_0(x) \sim \mathbb{N}(0,625(-\Delta+25I)^{-2})$ \cite{li2020fourier}. We utilize 1000 training and 100 testing samples to test the performance of the proposed framework on a spatial grid of 1024. Although the Burgers equation is a time-dependent differential equation, following the original problem statement in \cite{li2020fourier}, we aim to learn the solution operator $\mathcal{N}_{\Omega}: u_0(x) \mapsto u_1(x)$, i.e., the integral operator which maps the initial conditions to the solution at $T=1$s. 
The exploration space for the wavelet basis functions in the FWNO follows the following set \{$db$6, $coif$6, $bior$6.8, $rbio$6.8, $sym$6\}, where $db$ refers to Daubechies, $coif$ refers to Coiflet, $bior$ refers to Biorthogonal, $rbio$ refers to Reverse Biorthogonal, and $sym$ refers to Symlet wavelet family. The exploration space for the activation functions has the following set \{GeLU, ELU, Leaky ReLU, SELU, Sigmoid\}. 
Neural network $N_w$ is initialized with the db6 wavelet, and $N_a$ is initialized with the GeLU activation for each wavelet integral layer.

% \begin{table}[!ht]
%   \centering
%   \caption{Number of epochs required to train the Network Architectures considered}
%   \label{table:epochs}
%   \begin{tabular}{*{10}{c}}
%     \toprule
%     Architectures & \multicolumn{3}{c}{Number of Epochs} \\
%     \cmidrule(lr){2-4}
%     & Burgers & Darcy Flow & Navier-Stokes \\
%     \midrule
%     DeepONet & 500000& 100000& 100000 & \\ 
%     POD-DeepONet & 500000 & 100000& 100000& \\ 
%     FNO & 500 & 500& 500 &\\ 
%     MWT & 500 & 500& 500&\\ 
%     WNO & 500 & 1000& 800&\\ 
%     \bottomrule
%   \end{tabular}
% \end{table}

\begin{figure}[!ht]
    \centering
    \includegraphics[width=0.75\textwidth]{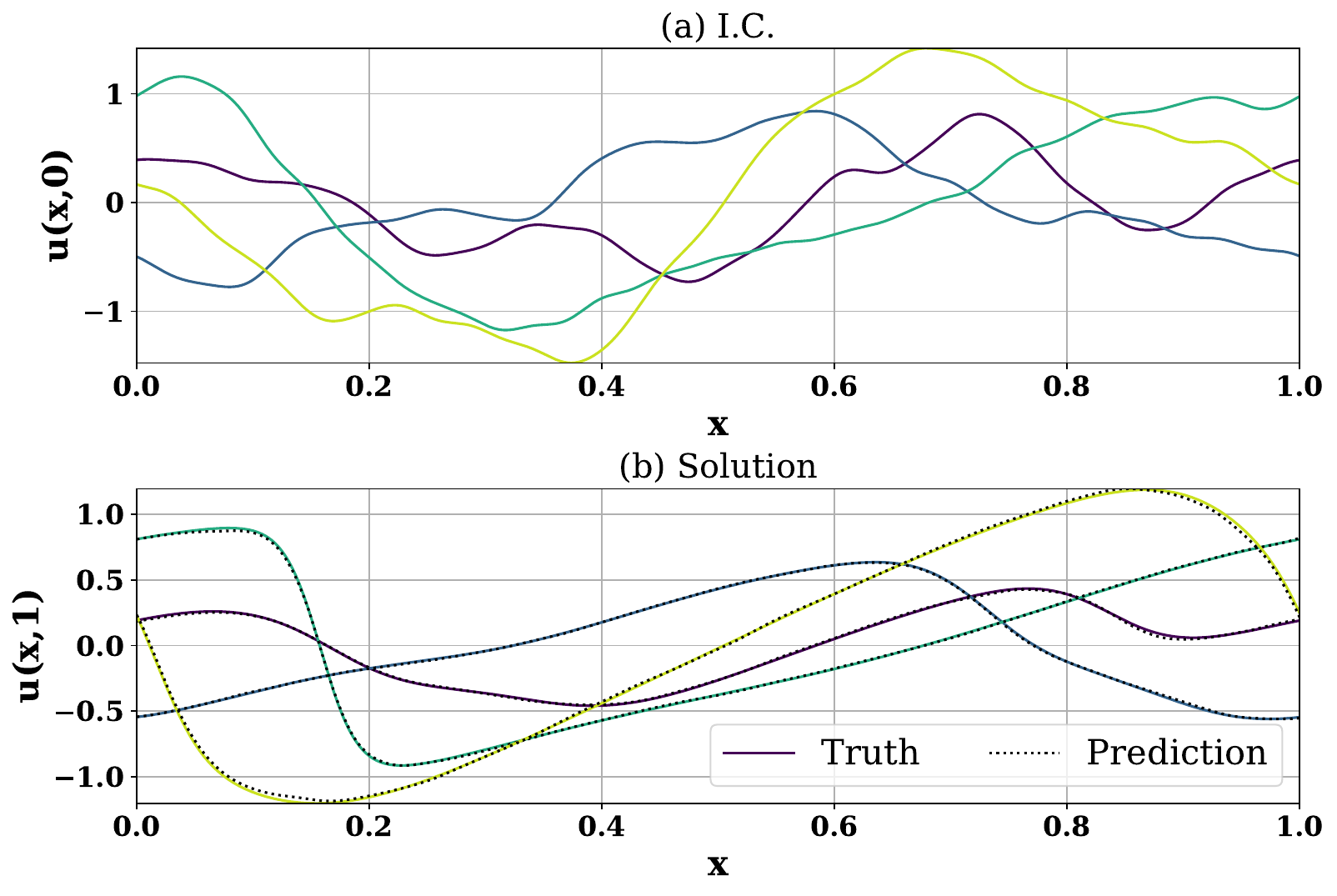}
    \caption{1D Burgers equation with periodic boundary conditions. (a) Four representative samples of initial conditions. (b) Ground truth vs. the prediction from best-performing WNO architecture sampled by FWNO at $T=1$s. The predictions suggest that the FWNO approximated solution operator approximates the true integral operator very closely. Differences between the truth and prediction are not easily discerned from the figure.}
    \label{fig:burgers}
\end{figure}

After the training, among the states sampled by the FWNO, we found the best-performing architecture to be constructed by the state \{$bior$6.8, GeLU, $db$6, GeLU, $rbio$6.8, GeLU, $db$6, ELU\}. This indicates that for the first wavelet integral layer, the wavelet basis and activation pair \{$bior$6.8, GeLU\} is the most probable pair that maximizes the reward and minimizes the flow discrepancy between the first two layers. Similarly, for the second, third, and fourth layers, the combinations \{$db$6, GeLU\}, \{$rbio$6.8, GeLU\}, and \{$db$6, ELU\} are found to be the most optimal pairs. 
The prediction results for four different representative initial conditions are illustrated in Fig. \ref{fig:burgers}. The mean relative $L^2$ error norm over the entire set of testing dataset is given in Table \ref{table:errors}. It is evident that the proposed FWNO clearly outperforms the vanilla WNO. A quantitative representation of the results obtained using the proposed approach is shown in Fig. \ref{fig:burgers}.
\begin{figure}[!ht]
    \centering
    \includegraphics[width=0.85\textwidth]{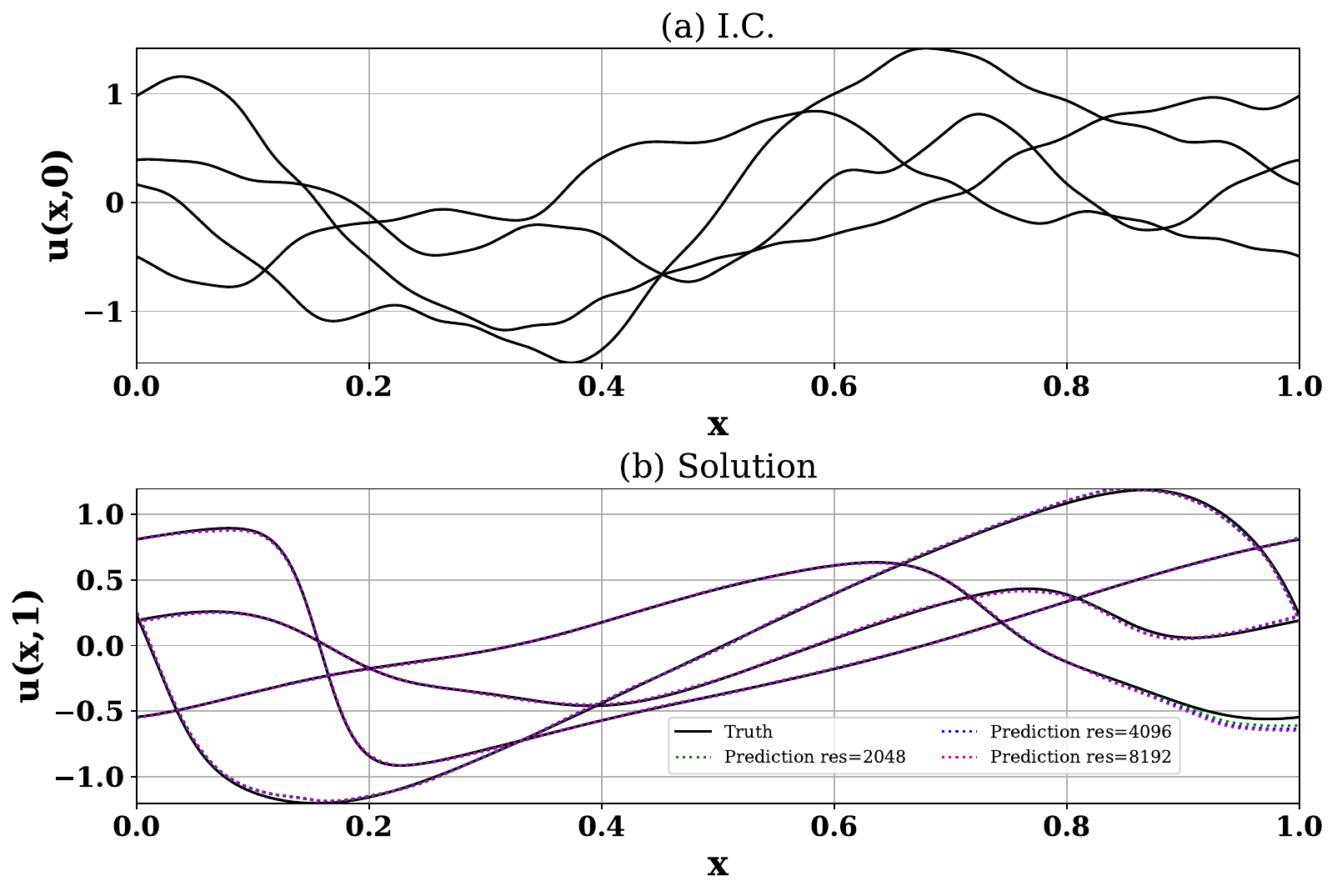}
    \caption{Zero-shot super-resolution on 1D Burgers equation. (a) Four representative samples of initial conditions from the test dataset. (b) Comparison of ground truth and predictions from the best-performing model at $T=1$s. The model is trained on a spatial resolution of 1024, while predictions are made on 2048, 4096, and 8192 resolutions. The predictions suggest the capability of the FWNO approximated solution operators to make zero-shot predictions on the higher-resolution datasets without fine-tuning.}
    \label{fig:burgers_1_super}
\end{figure}

In operator learning, a major point of interest lies in performance under higher resolution from what the operator is trained on during testing, also called super-resolution. Fig. \ref{fig:burgers_1_super} illustrates the experiments we conducted for different resolutions using the generated architecture for this example. It is clearly observed from Fig. \ref{fig:burgers_1_super} that despite being trained at a lower resolution, the generated architecture manages to produce highly accurate results on higher-resolution inputs as well.

\begin{figure}[!ht]
    \centering
    \includegraphics[width=\textwidth]{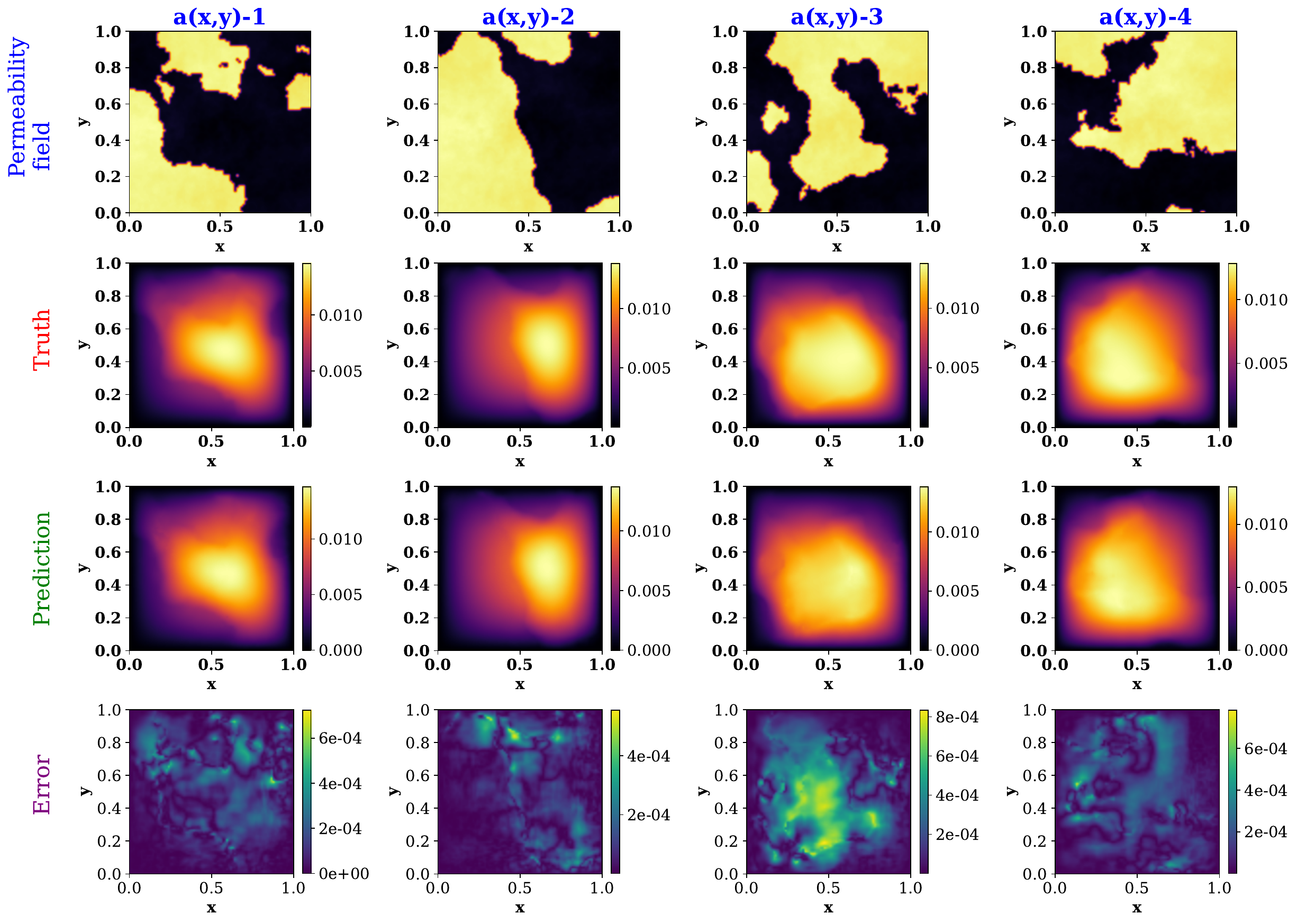}
    \caption{Time independent Darcy Flow in a rectangular domain. The plots show four representative samples of permeability fields, corresponding true pressure fields, and pressure field prediction, along with the absolute error plot for the best case among all states sampled by the FWNO. The figures show almost an exact match between the FWNO and the true solution.}
    \label{fig:Darcy}
\end{figure}
\subsection{Darcy's flow equation}
In the second problem, we consider the Darcy flow equation in a rectangular domain, which plays an important role in modeling the fluid flow through porous mediums. The partial differential description of the Darcy equation is given as,
\begin{equation}
    \begin{aligned}
        -\nabla \cdot \left(a(x,y)\nabla u(x,y)\right) &= f(x,y), \;\; x,y \in [0,1] \\
        u(x,y) &= 0, \;\; x,y \in \partial D
    \end{aligned}
\end{equation}
where $u(x,y)$ represents the pressure field, $a(x,y)$ represents the permeability field, $f(x,y)$ represents the source field, and $\partial D$ represents the boundary of the domain. The training data set is generated for different permeability fields $a(x,y) = \psi \mathbb{N}(0,-\Delta + 9I)^{-2}$, where $\psi: \mathbb{R} \mapsto \mathbb{R}$ is a pointwise push-forward operation that takes a value of 12 for the positive part of real line and 3 on the negative part  \cite{li2020fourier}. The aim in this example is to learn the solution operator $\mathcal{N}_{\Omega}: a(x,y) \mapsto u(x,y)$, i.e., the integral operator which maps the permeability fields to the pressure fields. 
The wavelet basis search space is taken as the set \{$db$4, $coif$6, $bior$6.8, $rbio$6.8, $sym$6\}, where the prefix of the bases are defined in the previous example. The search space of the activation operator is \{GeLU, ELU, Leaky ReLU, SELU, Sigmoid\}. Similar to the Burgers example, we use two feed-forward neural networks, $N_w$ and $N_a$, to choose the wavelet basis and activation operators from these sets.

% \begin{table}[!ht]
%     \centering
%     \caption{Mean $L_2$ relative error on testing set}
%     \label{table:errors}
%     \begin{adjustbox}{width=\textwidth}
%     \begin{tabular}{l*{10}{c}}
%         \toprule
%         PDE examples & \multicolumn{7}{c}{Neural operator architectures} \\
%         \cmidrule(lr){2-8}
%         & GNO & DeepONet & FNO & MWT & POD-DeepONet$^*$ & WNO & FWNO\\
%         \midrule
%         Burgers' Equation & 6.15\%&2.15\%&1.60\%& 0.19\%&1.94\%&1.75\%&1.44\%\\
%         Darcy Equation$^{**}$ & 3.46\%&2.98\%&1.08\%& 0.89\%&2.38\%&1.8\%&1.58\%\\
%         Darcy Equation (Triangular) & -&2.64\%&-& 0.87\%&1\%&0.88\%&0.59\%\\
%         Navier Stokes Equation & -&1.78\%&1.28\%& 0.63\%&1.36\%&3.43\%&2.35\%\\
%         \bottomrule
%     \end{tabular}
%     \end{adjustbox}
%     \footnotesize{$^{*}$POD-DeepONet is a version of vanilla DeepONet \cite{lu2022comprehensive}, which learns the solution operator from a space featured by proper orthogonal decomposition (POD). $^{**}$The GFlowNet was trained for WNO architectures of width 64, leading to better results from the simple WNO trained on a width of 128 in \cite{TRIPURA2023115783}. All WNO results reported here are based on a width of 64.}
% \end{table}
\begin{table}[!ht]
    \centering
    \caption{Mean relative $L_2$ prediction error on testing set}
    \label{table:errors}
    \begin{threeparttable}
    \begin{tabular}{lllll}
        \toprule
        PDE examples & Burgers' equation & Darcy equation & Darcy (Triangular) & Navier-Stokes equation \\
        \midrule
        WNO$^{a}$ & 1.75\% & $^*$1.8\% & 0.88\% & 3.43\% \\
        FWNO$^{b}$ & 1.44\% & 1.58\% & 0.59\% & 2.35\% \\
        \bottomrule
    \end{tabular}
    \begin{tablenotes}
       \item [a,b] WNO results are obtained on an uplifting dimension of 64. Similarly, the FWNO was trained for WNO architectures of width 64. $^*$However, FWNO provides better accuracy on the Darcy equation, where WNO was trained on an uplifting dimension of 128 \cite{TRIPURA2023115783}. 
    \end{tablenotes}
    \end{threeparttable}
\end{table}

Upon training, among the states sampled by the network, the best-performing network architecture for the Darcy flow equation is found to be constructed by the state \{$sym$6, GeLU, $db$4, GeLU, $sym$6, GeLU, $db$4, Leaky ReLU\}. We observe that only in the second wavelet integral layer, the wavelet basis and activation operator pair \{$db$4, GeLU\}, which was used in the vanilla WNO paper, obtains the highest probability among other states. In the other layers, new combinations with the wavelet basis $db$4 and Leaky ReLU are found to be optimal for maintaining flow between network layers.
The prediction results of pressure fields for four representative permeability fields are illustrated in Fig. \ref{fig:Darcy}. The associated relative $L_2$ error norms are provided in Table \ref{table:errors}. It is seen that our method outperforms the results obtained using vanilla WNO as reported in \cite{TRIPURA2023115783}.

\begin{figure}[!ht]
    \centering
    \includegraphics[width=\textwidth]{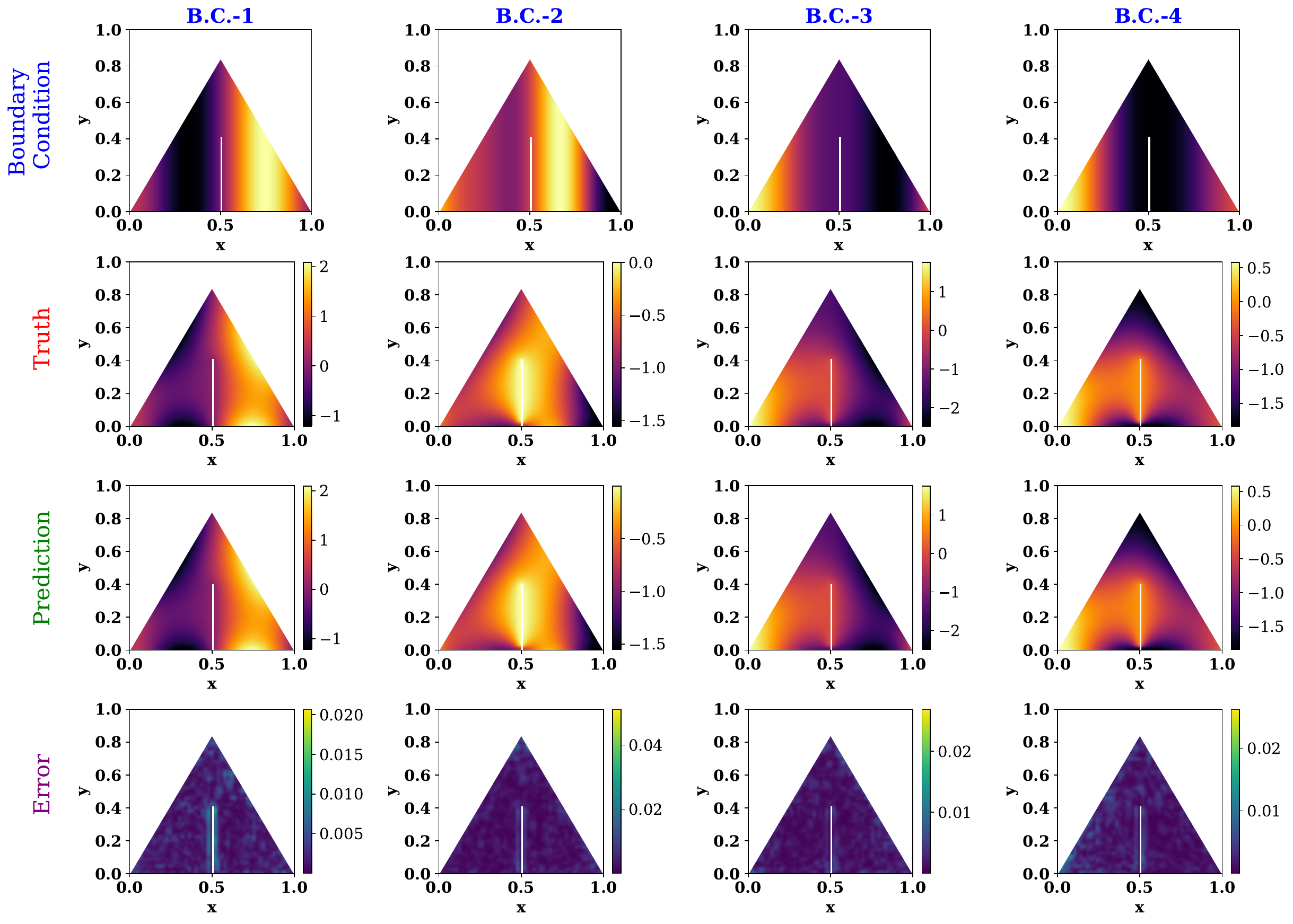}
    \caption{Darcy flow simulation in a triangular domain with a notch. The plots show four representative boundary conditions, the corresponding true pressure fields, and predicted pressure fields for the best case among all states sampled by the FWNO along the absolute error fields. The optimal solutions show an almost exact match with the ground truth, even for a complex geometry.}
    \label{fig:DarcyTri}
\end{figure}
\subsection{Darcy's flow equation in triangular geometry with a notch}
As a follow-up to the previous Darcy equation in the rectangular domain, we further consider the same partial differential equation with a triangular domain and an added notch in the flow medium. In this example, our aim is to learn the operator $\mathcal{N}_{\Omega}: u(x,y)\vert_{\partial D} \mapsto u(x,y)$, which maps the boundary conditions to the pressure field. Due to the presence of the notch and the triangular domain, predicting the pressure field from boundary conditions becomes difficult. 
Different boundary conditions for the training dataset are modeled as random fields with the radial basis function kernel as,
\begin{equation}
    \begin{aligned}
        u(x,y) \vert_{\partial D} & \sim GP\left(0, \kappa\left(x,y,x^{\prime},y^{\prime}\right)\right), \;\; x \in [0,1] \\
        \kappa(x,y,x^{\prime},y^{\prime}) &= \exp\left(-\left( \frac{(x-x')^2}{2l_x^2} + \frac{(y-y')^2}{2l_y^2} \right)\right) , 
    \end{aligned}
\end{equation}
where for data generation $l_x = l_y = 0.2$ is considered. The permeability field and the forcing function are kept equal to $a(x,y)=0.1$ and $f(x,y)=-1$, respectively \cite{lu2022comprehensive}. The search space for the wavelets and activation operators consists of the set \{$db$6, $coif$6, $bior$6.8, $rbio$6.8, $sym$6\} and \{GeLU, ELU, Leaky ReLU, SELU, Sigmoid\}, respectively. 
Among all states sampled, the best-performing state was \{$db$6, GeLU, $rbio$6.8, ELU, $db$6, ELU, $coif$6, GeLU\}. While the pair \{$db$6, GeLU\}, which was used in the vanilla WNO paper, has the highest probability in the first wavelet integral layer, the FWNO adapts the initial choices to the best possible combination as the training of FWNO progresses. The prediction results of pressure fields for four representative boundary conditions are illustrated in Fig. \ref{fig:DarcyTri}. The mean estimate of the relative $L^2$ prediction error over the testing dataset is provided in Table \ref{table:errors}. The predictions and absolute error plots in Fig. \ref{fig:DarcyTri} indicate an improvement of the vanilla WNO architecture, where all the wavelet integral layers use the same wavelet-activation pair. 

\begin{figure}[!ht]
    \centering
    \includegraphics[width=\textwidth]{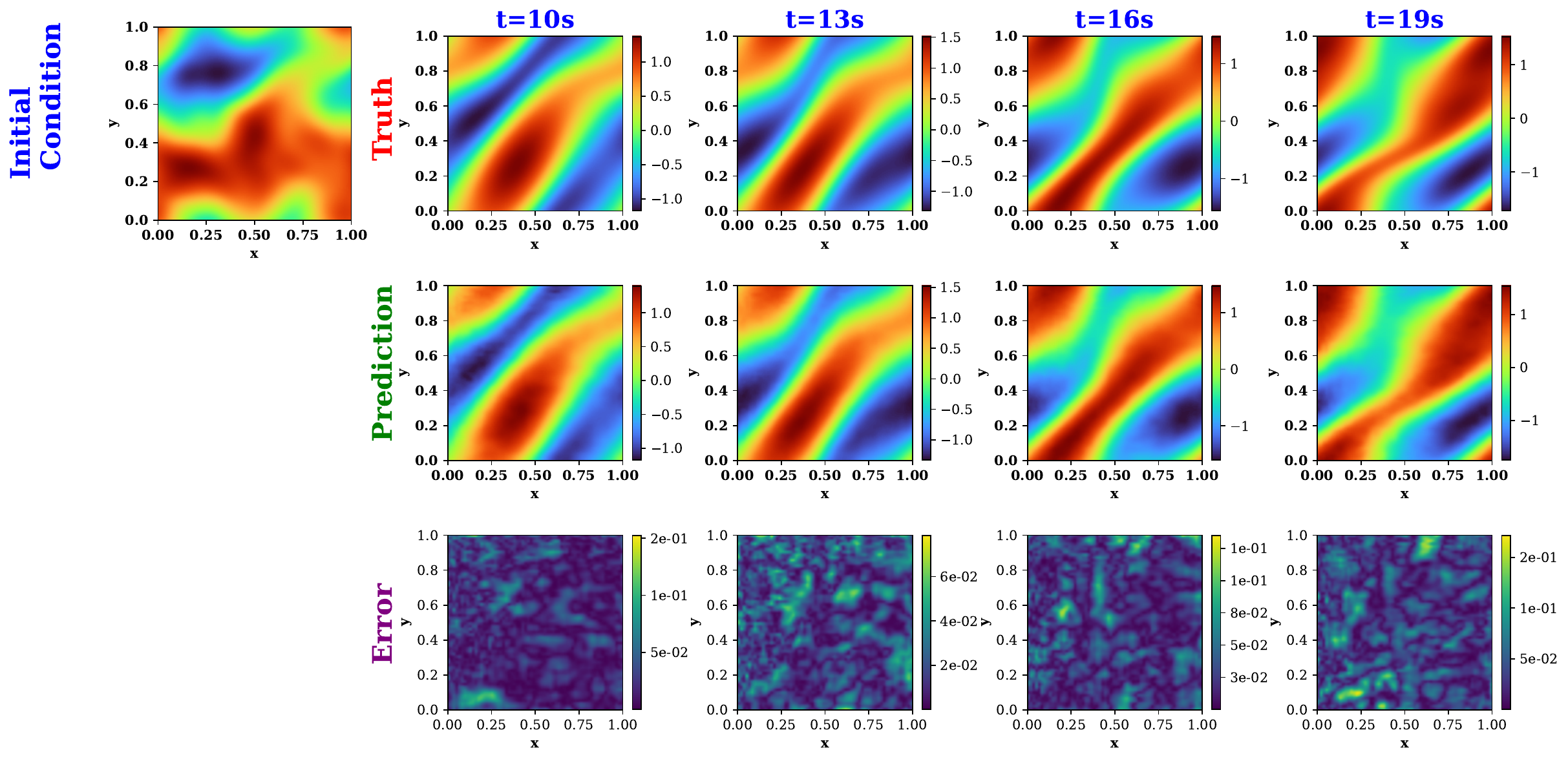}
    \caption{Incompressible Navier-Stokes equation with a spatial resolution of $64 \times 64$. The plots represent the initial vorticity field, the time evolution of the vorticity field at $t \in [11,20]$, the predicted vorticity field for the best case among all states sampled by the FWNO, and the error fields. The proposed framework shows a consistent performance over all the predicted time steps.}
    \label{fig:NS}
\end{figure}
\subsection{Navier-Stokes viscous fluid dynamics}
The Navier-Stokes equation is a nonlinear coupled second-order partial differential equation that plays a fundamental role in describing the dynamic behavior of viscous fluid flow. These equations have widespread applications, playing a crucial role in aerodynamics for aircraft design, numerical simulations of weather patterns, and the study of physiological phenomena like blood circulation in biological systems.
Due to the nonlinear divergence and diffusion terms, simulating the velocity fields of the Navier-Stokes equation becomes highly challenging. To that end, we consider the 2D incompressible Navier-Stokes equation in its vorticity form, given as,
\begin{equation}
    \begin{aligned}
        \frac{\partial \omega(x,y,t)}{\partial t} &= - \bm{u}(x, y, t) \cdot \nabla \omega(x, y, t) + \nu \Delta \omega(x, y, t) + f(x, y), \;\; & x,y \in [0,1], \; t \in [0,T] \\
        \nabla \cdot \bm{u}(x,y,t) &= 0, \;\; & x,y \in [0,1], \; t \in [0,T] \\
        \omega(x,y,0) &= \omega_0(x, y), \;\; & x,y \in [0,1]
    \end{aligned}
\end{equation}
where $\nu \in \mathbb{R}_{>0}$ is the positive viscosity coefficient of the viscous flow, $u(x,y,t)$ is the velocity of the viscous flow, $\omega(x,y,t)$ is the vorticity field, and the $f(x,y)$ is the source field. For training data generation, the viscosity is taken as $\nu=10^{-3}$, and the force field is defined as $f(x,y)=0.1(\operatorname{sin}(2\pi(x+y)) + \operatorname{cos}(2\pi(x+y)))$. The datasets are generated for different initial vorticity fields, which are modeled as random Gaussian fields as $\omega_0(x,y) = \mathbb{N}(0,7^{1.5}(-\Delta + 49I)^{-2.5})$ \cite{li2020fourier}. The vorticity fields are obtained at resolutions $64 \times 64$. The aim is to learn a time-dependent operator $\mathcal{N}_{\Omega}: \omega \vert_{[0,1]^2 \times [1,10]} \mapsto \omega \vert_{[0,1]^2 \times [11,20]}$, that maps the vorticity fields at first ten-time steps to next ten time steps for arbitrary initial vorticity fields. 
For constructing the best architecture, we construct the search space of wavelet basis as \{$db$4, $coif$6, $bior$6.8, $rbio$6.8, $sym$6\} and \{GeLU, ELU, Mish, SELU, Sigmoid\} for the activation operators. 
% Here, we introduce the Mish activation for this example in the activation function search space. 
Each wavelet integral layer of the WNO operator is initialized on a $db$4 wavelet basis and with a GeLU activation operator for each layer of the WNO. 

Among all the sampled states, the best-performing state is found to be \{$db$4, GeLU, $sym$6, GeLU, $rbio$6.8, GeLU, $db$4, GeLU\}. While the states with the highest probability are dominated by the Daubechies wavelets and GeLU activation function, the best-performing state is participated by three different wavelet basis functions. 
These prediction results of the vorticity fields from the proposed FWNO architecture are illustrated in Fig. \ref{fig:NS}. The mean estimate of the relative $L^2$ error norm over the test dataset is provided in Table \ref{table:errors}. In Fig. \ref{fig:NS}, we observe a similar performance as the previous examples. The architecture constructed by the best-performing state maintains a consistent accuracy over all the prediction time steps. The mean error values in Table \ref{table:errors} are also evidence that the proposed framework outperforms the vanilla WNO architecture.

\section{Conclusions}\label{sec:conclusion}
% What was done here (3-4 sentences)? \\
% What did we achieve? \\
% What is novel? 
% What impact these results will have? \\
% What next? \\
In this article, we proposed a generative flow-based neural architecture search algorithm for neural operators. The neural search architecture is implemented on the recently proposed WNO, thereby introducing the flow-induced wavelet neural operator (FWNO). The resulting framework generates its own architecture by generating the sequence of the best-performing pairs of wavelet basis and activation operators in the vanilla WNO. The sequence is generated by learning stochastic policies through simple feed-forward neural networks and can also be extended to other network hyperparameters. 
Compared to the computationally involved grid search-reliant hyperparameters tuning algorithm, the FWNO employs a flow cum reward-based strategy, where the flow discrepancy between two states is reduced, and the reward from the terminal state is increased to learn a probabilistic policy to generate the most appropriate sequence of wavelet basis and activation operators.  
Once the training succeeds, the learned policies sample the states (wavelet bases and activation operators) in proportion to the reward at the terminal state so as to deliver the best-performing architectures among the diverse set of possible sequences. At the terminal state, i.e., the WNO model, we return the exponent of the negative of the prediction loss on the validation dataset. 

On a broader level, the FWNO inherits all the properties of the WNO and learns a discretization invariant functional mapping between infinite-dimensional function spaces using the frequency-space localization property wavelet decomposition. Therefore, with a single training, it can deployed for reliable prediction of solutions on a different grid size for a family of parametric PDEs. 
We exemplified the efficacy of the proposed framework on four operator learning benchmarks. In each example, we found reductions in error when compared with the vanilla WNO. In the case of Burger's equation, our method reduces the relative $L_2$ error from 1.75\% to 1.44\%. On the Darcy flow equation, the error reduces from 1.8\% to 1.58\% on a rectangular grid while it drops from 0.88\% to 0.59\% on a triangular grid, showcasing it has preserved all the properties of vanilla WNO. For the incompressible Navier Stokes equation, FWNO reduces errors significantly from 3.43\% to 2.35\%, with over an entire percentage change when compared with the vanilla WNO.

Before concluding the discussion, we summarise the contribution of this study in the following points, 
\begin{itemize}
    \item We have illustrated a generative flow-based architectural search algorithm for neural operators. By setting the flow consistency between integral layers of neural operators, the proposed framework learns policies to sample a best-performing set of network hyperparameters in proportion to the positive rewards of the actions taken. 
    \item We have showcased the efficacy of the framework on the WNO, which contains a multitude of hyperparameters like the wavelet basis and activation operator. Instead of selecting a fixed set of wavelets and activation operators like in vanilla WNO, the proposed framework selects layers-specific pairs of wavelets and activation operators. 
\end{itemize}
To conclude, we note that while the implementation of the flow-based strategy is limited to the wavelet neural operator in this paper, the method proposed can be seamlessly integrated with other neural operators and neural networks as well. 
Future work on the same could incorporate other hyper-parameters, such as the number of convolution layers and the number of layers for wavelet decomposition.

\section*{Code and data availability}
Upon acceptance, all the source codes to reproduce the results in this study will be made available to the public on GitHub by the corresponding author.

% \bibliographystyle{unsrt}
% \bibliography{references}  

%%% Uncomment this line and comment out the ``thebibliography'' section below to use the external .bib file (using bibtex).



% \end{thebibliography}

\end{document}